\documentclass[10pt,twocolumn,letterpaper]{article}

\usepackage[pagenumbers]{cvpr} 

\usepackage{times}
\usepackage{epsfig}
\usepackage{graphicx}
\usepackage{amsmath}
\usepackage{amssymb}
\usepackage{multicol}
\usepackage{multirow}
\usepackage{rotating}
\usepackage{booktabs}
\usepackage{colortbl}
\usepackage{bm}
\usepackage{wasysym}
\usepackage{utfsym}
\usepackage{wrapfig}

\definecolor{cvprblue}{rgb}{0.21,0.49,0.74}
\definecolor{mygray}{gray}{.9}
\usepackage[pagebackref,breaklinks,colorlinks,allcolors=cvprblue]{hyperref}

\def\paperID{6118} % *** Enter the Paper ID here
\def\confName{CVPR}
\def\confYear{2025}

\title{BVINet: Unlocking Blind Video Inpainting with Zero Annotations}
\author{
Zhiliang Wu,~~
Kerui Chen,~~
Kun Li,~~
Hehe Fan,~~
Yi Yang\\%$^{2,3}$\And\\
\\
ReLER Lab, CCAI, Zhejiang University\\
}

\begin{document}
\maketitle
\begin{abstract}
Video inpainting aims to fill in corrupted regions of the video with plausible contents.
Existing methods generally assume that the locations of corrupted regions are known, focusing primarily on the “how to inpaint”. 
This reliance necessitates manual annotation of the corrupted regions using binary masks to indicate “where to inpaint”. 
However, the annotation of these masks is labor-intensive and expensive, limiting the practicality of current methods.
In this paper, we expect to relax this assumption by defining a new blind video inpainting setting, enabling the networks to learn the mapping from corrupted video to inpainted result directly, eliminating the need of corrupted region annotations.
Specifically, we propose an end-to-end blind video inpainting network (BVINet) to address both “where to inpaint” and “how to inpaint” simultaneously.
On the one hand, BVINet can predict the masks of corrupted regions by detecting semantic-discontinuous regions of the frame and utilizing temporal consistency prior of the video.
On the other hand, the predicted masks are incorporated into the BVINet, allowing it to capture valid context information from uncorrupted regions to fill in corrupted ones.
Besides, we introduce a consistency loss to regularize the training parameters of BVINet.
In this way, mask prediction and video completion mutually constrain each other, thereby maximizing the overall performance of the trained model.
Furthermore, we customize a dataset consisting of synthetic corrupted videos, real-world corrupted videos, and their corresponding completed videos.
This dataset serves as a valuable resource for advancing blind video inpainting research.
Extensive experimental results demonstrate the effectiveness and superiority of our method.
\end{abstract}    
\begin{figure}[tb]
  \centering
  \includegraphics[scale=0.65]{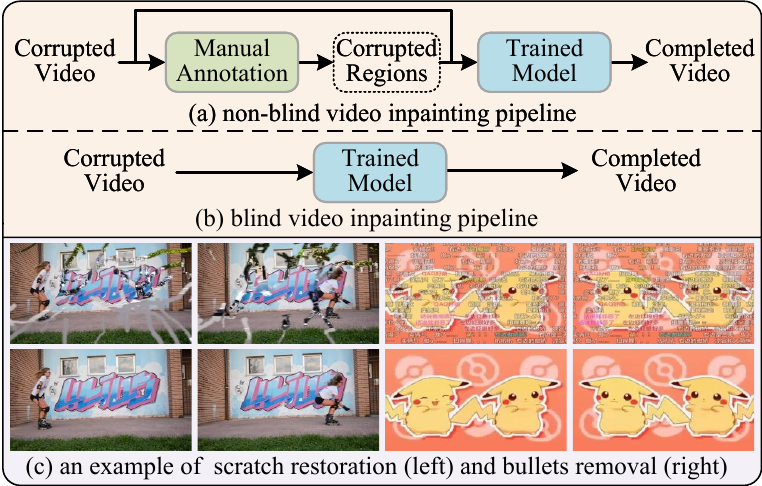}
  \vspace{-0.5cm}
  \caption{ Fig.(a) shows the general pipeline of existing non-blind video inpainting. 
  Such pipeline require manual annotation of corrupted regions of each frame, limiting its application scope. 
  In this paper, we formulate a new task: blind video inpainting, which can directly learn a mapping from corrupted video to inpainted result without any corrupted region annotation (Fig.(b)).
  Fig.(c) shows an example of our blind video inpainting method in scratch restoration and bullet removal.}
  \label{Fig1}
  \vspace{-0.5cm}
\end{figure}
\section{Introduction}
\label{Introduction}
Video inpainting is a fundamental visual restoration task in computer vision, aiming to fill corrupted regions in videos with plausible and coherent contents~\cite{Wu_2023_CVPR,9967838,Wu_2023_CVPR1}. 
Existing video inpainting methods~\cite{cai2022devit,liu2021fuseformer,Ren_2022_CVPR,yan2020sttn,zhang2022flow,cherel2023infusion,yu2023deficiency,Wu_2023_CVPR,Zheng_2023_ICCV} 
typically take corrupted videos along with masks indicating corrupted regions as input, and generate corrupted contents based on spatial texture and temporal information of valid (uncorrupted) regions.
This means that these methods require manual annotation of corrupted regions using binary masks to indicate “where to inpaint”.
We refer to this case as a \emph{non-blind setting}.
Nevertheless, accurate mask annotations are not available in many scenarios.
On the one hand, the boundary between corrupted regions and valid regions is often blurred, making it unrealistic to accurately annotate corrupted regions. 
On the other hand, manual annotation is labor-intensive and resource-consuming, especially for corrupted videos with high frame rates and high resolutions.

In this paper, we formulate a novel task: \textbf{\emph{blind video inpainting}}, where a mapping from corrupted videos to inpainted results is directly learned without any corrupted region annotation.
Unlike the non-blind setting, which only focuses on “how to inpaint”, our proposed blind video inpainting setting requires considering both “where to inpaint” and “how to inpaint” simultaneously. 
This blind video inpainting setting is more effective in real-world applications.
In fact, the “corruption” faced by video inpainting can be divided into two types.
The first type of “corruption” is not present in the original video.
This type of “corruption” is caused by external factors and destroys the original structure of the video, such as scratches~\cite{chang2019free,wu2023flow}, watermarks~\cite{9967838,10222097}, and bullets~\cite{Kim_2019_CVPR}, etc.
Another type of “corruption” exists in the original video itself, such as undesired objects and occlusions~\cite{9010390,10447901}. 
In this paper, we mainly focus on the first type of “corruption”.

For blind video inpainting task, a naive solution is to process corrupted video frame by frame using existing blind image inpainting methods~\cite{schmalfuss2022blind,wang2020vcnet,10147235}. 
However, such a solution will neglect the motion continuity between frames, resulting in flicker artifacts in the completed video.
In this paper, we propose an end-to-end blind video inpainting framework consisting of a mask prediction network and a video completion network.
The former aims to estimate the corrupted regions that need to be completed, while the latter focuses on completing the corrupted video.
Specifically, we first develop a mask prediction network to predict the corrupted regions of whole video by detecting semantic-discontinuous regions of the frame and utilizing temporal consistency prior of the video.
Then, we design a video completion network to perceive valid context information from uncorrupted regions using predicted mask to generate corrupted contents.
To precisely capture the accurate correspondence between the mask prediction network and video completion network, we introduce a consistency loss to regularize their parameters, enabling mutual constraint and maximizing the overall model performance.

Furthermore, existing video inpainting datasets~\cite{lee2019cpnet,chang2019free,yan2020sttn,Kang2022ErrorCF} usually introduce specific prior knowledge during the construction process, such as corrupted content, clear border, and fixed shape.
These priors make corrupted regions easily distinguishable from natural video frame by the mask prediction network, failing to realistically simulate the complex real-world scenarios in blind video inpainting.
In this paper, we exploit free-form strokes~\cite{chang2019free} as corrupted regions, and fill these regions with natural images as corrupted contents. 
Such operations allow us to construct the corrupted contents with variable shapes, complex motions and faithful texture.
To avoid the introducing border priors, we also utilize iterative Gaussian smoothing~\cite{wang2018image} to extend the edges of corrupted regions, making these boundaries difficult to distinguish.
Such strategy can enforce the network to infer the corrupted regions by the semantic context of the video frame rather than merely fitting to the training data.
Further, to enhance the generalization ability of our model in practical applications, we collect a bullet removal dataset consisting of 1,250 video clips. 
Extensive experimental results demonstrate that our method can achieve comparable performance to non-blind baselines. 
An example result of our method in scratch restoration and bullets removal is shown in Fig.\ref{Fig1}(c).

To sum up, our contributions are summarized as follows:
\begin{itemize}

\item A novel blind video inpainting task is formulated. It directly learns a mapping from corrupted videos to inpainted results without any corrupted region annotation. To the best of our knowledge, this is the first blind inpainting work in the field of video inpainting.
  
\item An end-to-end blind video inpainting framework consisting of a mask prediction network and a video completion network is designed, and a consistency loss is introduced to regularize the training parameters of this framework.
  
\item A dataset suitable for blind video inpainting task is customized. This dataset consists of $2,400$ synthesized video clips and $1,250$ real-world video clips, and will be published to facilitate subsequent research. 
\end{itemize}

\section{Related Works}
\label{Related Work}

\noindent\textbf{Non-Blind Video Inpainting.}
% \subsection{Video Inpainting}
% With the rapid development of deep learning, video inpainting has made great progress. 
The video inpainting methods can be roughly divided into three lines: 3D convolution-based~\cite{chang2019free,Kim_2019_CVPR,9558783}, flow-based~\cite{Gao-ECCV-FGVC,Kang2022ErrorCF,Ke2021OcclusionAwareVO,li2022towards,xu2019deep,Zhang_2022_CVPR,zou2020progressive}, and attention-based methods~\cite{cai2022devit,lee2019cpnet,Li2020ShortTermAL,liu2021fuseformer,Ren_2022_CVPR,9010390,srinivasan2021spatial,yan2020sttn,zhang2022flow}. 

%\noindent\textbf{3D convolution-based methods.}
The methods~\cite{chang2019free,Kim_2019_CVPR,9558783} based on 3D convolution usually reconstruct the corrupted contents by directly aggregating complementary information in a local temporal window through 3D temporal convolution. 
%For example,
%Wang et al.~\cite{wang2018video} proposed the first deep learning-based video inpainting network using a 3D encoder-decoder network.
%Further,
%Kim et al.~\cite{Kim_2019_CVPR} aggregated the temporal information of the neighbor frames into missing regions of the target frame by a recurrent 3D-2D feed-forward network.
Nevertheless, they often yield temporally inconsistent completed results due to the limited temporal receptive fields.
%\noindent\textbf{Flow-based methods.} 
The methods~\cite{Gao-ECCV-FGVC,Kang2022ErrorCF,Ke2021OcclusionAwareVO,li2022towards,xu2019deep,Zhang_2022_CVPR,zou2020progressive} based on optical flow treat the video inpainting as a pixel propagation problem. 
Generally, they first introduce a deep flow completion network to complete the optical flow, and then utilize the completed flow to guide the valid pixels into the corrupted regions. 
%For instance, 
%Xu et al.~\cite{xu2019deep} used the flow field completed by a coarse-to-fine deep flow completion network to capture the correspondence between the valid regions and the corrupted regions, and guide relevant pixels into the corrupted regions. 
%Based on this, 
%Gao et al.~\cite{Gao-ECCV-FGVC} further improved the performance of video inpainting by explicitly completing the flow edges. 
%Zou et al.~\cite{zou2020progressive} corrected the spatial misalignment in the temporal feature propagation stage by the completed optical flow. 
However, these methods fail to capture the visible contents of long-distance frames, thus reducing the inpainting performance in the scene of large objects and slowly moving objects.

%\noindent\textbf{Attention-based methods.} 
Due to its outstanding long-range modeling capacity, attention-based methods, especially transformer-based methods, have shed light on the video inpainting community. 
These methods~\cite{cai2022devit,lee2019cpnet,Li2020ShortTermAL,liu2021fuseformer,Ren_2022_CVPR,9010390,srinivasan2021spatial,yan2020sttn,zhang2022flow} first find the most relevant pixels in the video frame with the corrupted regions by the attention module, and then aggregate them to complete the video frame. 
%For example,
%Zeng et al.~\cite{yan2020sttn} ﬁlled the missing regions of multi-frames simultaneously by learning a spatial-temporal transformer network. 
%Further, Liu et al.~\cite{liu2021fuseformer} improved edge details of missing contents by novel soft split and soft composition operations.
Although the existing video inpainting methods have shown promising results, 
%they usually assume that the corrupted regions of the video are known, 
they usually need to elaborately annotate the corrupted regions of each frame in the video,
limiting its application scope.
Unlike these approaches, we propose a blind video inpainting network in this paper, which can automatically identify and complete the corrupted regions in the video.

\begin{figure*}[tb]
\centering%height=3.0cm,width=15.5cm
\includegraphics[scale=0.66]{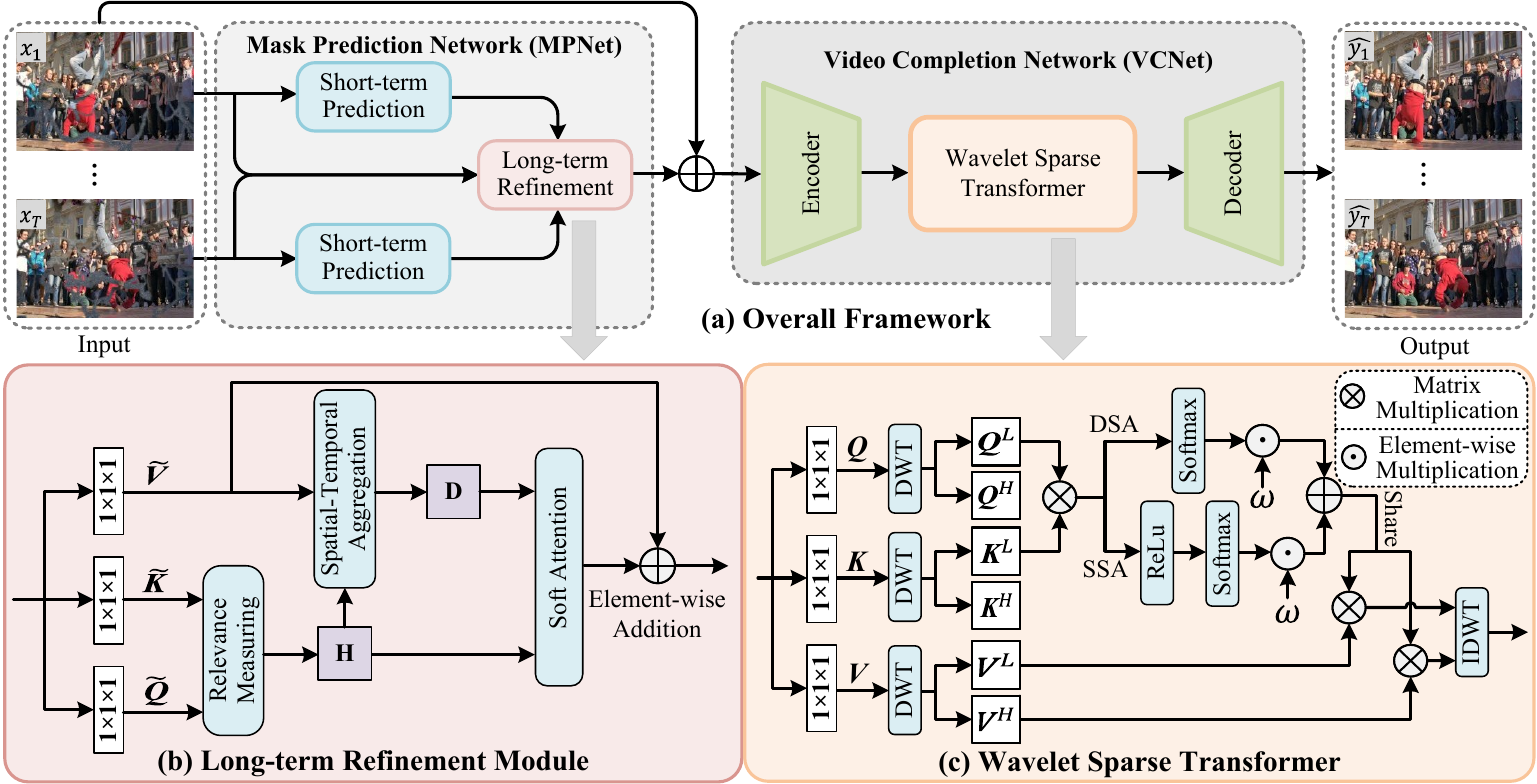}
\vspace{-0.15cm}
\caption{\textbf{The overview of the proposed blind video inpainting framework}. Our framework are composed of a mask prediction network (MPNet) and a video completion network (VCNet). The former aims to predict the masks of corrupted regions by detecting semantic-discontinuous regions of the frame and utilizing temporal consistency prior of the video, while the latter perceive valid context information from uncorrupted regions using predicted mask to generate corrupted contents.
}
\label{Fig_KT}
\vspace{-0.5cm}
\end{figure*}

\noindent\textbf{Blind Image Inpainting.}
% \subsection{Blind Image Inpainting}
% \subsection{Image Inpainting}
In contrast to video inpainting, image inpainting solely requires consideration of the spatial consistency of the inpainted results. In the few years, the success of deep learning has brought new opportunities to many vision tasks, which promoted the development of a large number of deep learning-based image inpainting methods~\cite{shamsolmoali2023transinpaint,dong2022incremental,liu2022reduce,li2022misf,cao2022learning}. 
As a sub-task of image inpainting, blind image inpainting~\cite{wang2020vcnet,zhao2022transcnn,li2024semid,li2023decontamination,10147235} has been preliminarily explored. 
For example,
Nian et al.~\cite{cai2017blind} proposed a novel blind inpainting method based on a fully convolutional neural network. 
Liu et al.~\cite{BII} designed a deep CNN to directly restore a clear image from a corrupted input. However, these blind inpainting work assumes that the corrupted regions are filled with constant values or Gaussian noise, which may be problematic when corrupted regions contain unknown content. To improve the applicability, Wang et al.~\cite{wang2020vcnet} relaxed this assumption and proposed a two-stage visual consistency network. 
%Jenny et al.~\cite{schmalfuss2022blind} improved inpainting quality by integrating theoretically founded concepts from transform domain methods and sparse approximations into a CNN-based approach. 
Compared with blind image inpainting, blind video inpainting presents an additional challenge in preserving temporal consistency. Naively applying blind image inpainting algorithms on individual video frame to fill corrupted regions will lose inter-frame motion continuity, resulting in flicker artifacts. Inspired by the success of deep learning in blind image inpainting task, we propose the first deep blind video inpainting model in this paper, which provides a strong benchmark for subsequent research.

\section{Proposed Method}
\label{Method}
\subsection{Problem Formulation}
\label{Problem Formulation}
Given a corrupted video sequence $\textbf{\emph{X}}=\{\textbf{\emph{x}}_1,\textbf{\emph{x}}_2,\dots,\textbf{\emph{x}}_T\}$ with sequence length $T$. 
The corrupted regions of video sequence $X$ are denoted as a binary mask $\textbf{\emph{M}}=\{\textbf{\emph{m}}_1,\textbf{\emph{m}}_2, \dots,\\
\textbf{\emph{m}}_T\}$. 
For each mask $\textbf{\emph{m}}_i$, “0” indicates that corresponding pixel is corrupted, and “1” denotes the valid regions.  
The goal of video inpainting is to generate a complete video sequence $\widehat{\textbf{\emph{Y}}}=\{\widehat{\textbf{\emph{y}}}_1,\widehat{\textbf{\emph{y}}}_2 ,\dots,\widehat{\textbf{\emph{y}}}_T\}$, which should be spatially and temporally consistent with the ground truth $\textbf{\emph{Y}}=\{\textbf{\emph{y}}_1, \textbf{\emph{y}}_2,\dots, \textbf{\emph{y}}_T\}$. Formally, existing video inpainting methods~\cite{cai2022devit,liu2021fuseformer,Ren_2022_CVPR,yan2020sttn,zhang2022flow} aim
to model the conditional distribution $p(\widehat{\textbf{\emph{Y}}}|\textbf{\emph{X}},\textbf{\emph{M}})$ by training a deep neural network $\mathcal{D}$, \emph{i.e.}, $\widehat{\textbf{\emph{Y}}}=\mathcal{D}(\textbf{\emph{X}},\textbf{\emph{M}})$. 
Nevertheless, these methods are only applicable
when the corrupted region mask $\textbf{\emph{m}}_i$ of each video frame $\textbf{\emph{x}}_i$ is available. 
In practice,
accurate mask of corrupted regions is difficult to manually annotate due to the blurred boundary between the corrupted regions and the valid regions.
Furthermore, collecting these manual annotation mask is not only labor-intensive and time-consuming,
but also often subjective and error-prone.

Different from existing video inpainting~\cite{cai2022devit,liu2021fuseformer,Ren_2022_CVPR,yan2020sttn,zhang2022flow}, 
we focus on video inpainting without any manual annotation mask of corrupted regions,
called \emph{\textbf{blind video inpainting}}.
This task aims to learning a direct mapping {$\mathcal{G}$} from the corrupted video $\textbf{\emph{X}}$ to the completed video $\widehat{\textbf{\emph{Y}}}$ 
\emph{i.e.}, $\widehat{\textbf{\emph{Y}}}=\mathcal{G}(\textbf{\emph{X}})$.
It means that the mapping $\mathcal{G}$ needs not only to automatically locate the corrupted regions, but also complete them.
Towards this end,
we decompose the blind video inpainting task into two sub-tasks: \emph{mask prediction} and \emph{video completion}.
The former aims to estimate the corrupted regions of video, indicating “where to inpaint”, while the latter focus to complete the corrupted regions of video, addressing “how to inpaint”.
The dual sub-tasks are defined as follows:
\begin{itemize}
    \item \textbf{mask prediction}: 
    given a corrupted video sequence $\textbf{\emph{X}}$, it learns the mapping $\bm\Phi$ to predict the corrupted region masks $\textbf{\emph{M}}$, \emph{i.e.}, $\textbf{\emph{M}} = \bm \Phi(\textbf{\emph{X}})$;
    
    \item \textbf{video completion}:
    based on the predicted region masks $\textbf{\emph{M}}$,
    it learns the mapping $\bm\Psi$ to generate the completed video sequence $\widehat{\textbf{\emph{Y}}}$, \emph{i.e.}, 
    $\widehat{\textbf{\emph{Y}}} = \bm \Psi(\textbf{\emph{X}},\textbf{\emph{M}})$.
\end{itemize}

\subsection{Network Design}
\label{Network Design}

In this paper, we design an end-to-end trainable framework to tackle blind video inpainting task. 
As shown in Fig.\ref{Fig_KT}(a), our framework consists of a mask prediction network and a video completion network.
The former aims to predict the corrupted regions of entire video by detecting semantic-discontinuous regions of the frame and utilizing temporal consistency prior of the video, while the latter perceive valid context information from uncorrupted regions using predicted masks to generate corrupted contents.

\subsubsection{Mask Prediction Network}
As shown in Fig.\ref{Fig_KT}(a), 
our mask prediction network sub-network consists of a short-term prediction module and a long-term refinement module.
The short-term prediction module takes the individual video frame $\textbf{\emph{x}}_i$ as input to predict the binary mask $\textbf{\emph{m}}_i^s$ of visual inconsistency regions, thus providing initial input for the long-term refinement module.
The long-term refinement module is used to refine $T$ predicted masks
$\textbf{\emph{M}}^s = \{\textbf{\emph{m}}_1^s, \textbf{\emph{m}}_2^s, \dots, \textbf{\emph{m}}_T^s\}$
using the temporal consistency priors of videos.

\noindent\textbf{Short-Term Prediction Module}.
The short-term prediction module is designed as an encoder-decoder structure. 
Specifically, the encoder consists of three stages, which generate feature maps at three different scales.
All stages share a similar structure, which consists of multiple convolutional layers and residual blocks with ReLU activation functions. 
In contrast, the decoder is used to decode the learned features into the binary masks, and its architecture is similar to that of the encoder.
Furthermore, researchers have demonstrated that noise can significantly reduce image segmentation performance~\cite{9156335,9508165}. To enhance the noise immunity of CNN, Li et al.~\cite{9156335,9508165} replaced the max-pooling, strided-convolution, and average-pooling in CNN with DWT, thereby achieving higher image segmentation accuracy. 
Motivated by such observation, we employ DWT to perform the down-sampling operation in the short-term prediction module. 
In this way, the short-term prediction module can obtain a more accurate mask of corrupted regions.
Formally, the entire mask prediction process can be expressed as:
\begin{equation}
\textbf{\emph{m}}_i^s=STP(\textbf{\emph{x}}_i),
\label{STPM}
\end{equation}
where 
$STP$ denotes short-term prediction module.
$\textbf{\emph{m}}_i^s$ is the predicted mask of the $i$-th video frame $\textbf{\emph{x}}_i$.

\noindent\textbf{Long-Term Refinement Module}.
The long-term refinement module consists of three main parts: encoder, sequence-to-sequence transformer, and decoder, 
where the encoder and decoder have the same backbone as the encoder and decoder of the short-term refinement module.
Sequence-to-sequence transformer is the core component of the long-term refinement module. 
It aims to refine $\textbf{\emph{M}}^s = \{\textbf{\emph{m}}_1^s, \textbf{\emph{m}}_2^s, \dots, \textbf{\emph{m}}_T^s\}$ using the temporal consistency priors.

As shown in Fig.\ref{Fig_KT}(b), for deep features $\textbf{\emph{E}}=\{\textbf{\emph{e}}_1, \textbf{\emph{e}}_2, \cdots,\textbf{\emph{e}}_T\}$ extracted with encoder through cascading the corrupted video sequence $\textbf{\emph{X}}$ and mask $\textbf{\emph{M}}^s$, 
we first project them
into query (${\widetilde{\textbf{\emph{Q}}}}$), key (${\widetilde{\textbf{\emph{K}}}}$) and value (${\widetilde{\textbf{\emph{V}}}}$) using a $1\times1\times1$ convolutional layer.
After obtaining ${\textbf{\emph{Q}}}$, ${\textbf{\emph{K}}}$, and ${\textbf{\emph{V}}}$, they are split into $N$ groups \{$\widetilde{ \textbf{\emph{Q}}}_n, \widetilde{\textbf{\emph{K}}}_n, \widetilde{\textbf{\emph{V}}}_n\} \in \mathbb{R}^{T\times{H}\times{W}\times{\frac{C}{N}}}$
along the channel dimension, 
where $n \in \{1,2,\cdots,N\}$. 

Then, we measure the relevance between $\widetilde{\textbf{\emph{Q}}}_n$ and $\widetilde{\textbf{\emph{K}}}_n$ to extract the spatial-temporal relationship between video frames.
However, the positions of corrupted regions in adjacent frames are similar~\cite{gu2020pyramid,ji2021progressively}. Therefore, we use the surrounding neighborhood obtained from a response window to compute a spatial-temporal affinity matrix ${\textbf{\emph{H}}}_n$ of the target pixel, instead of computing the response between a query position and the features at all positions. 
Such a relevance measurement captures more relevance related to the target regions within $T$ frames. 

Having the affinity matrix ${\textbf{\emph{H}}}_i$, we calculate the spatial-temporally aggregated features ${\textbf{\emph{D}}}_n$ within the surrounding neighborhood by matrix multiplication between the response window of $\widetilde{\textbf{\emph{V}}}_n$ and the affinity matrix ${\textbf{\emph{H}}}_n$. 
Finally, a soft-attention block is used to synthesize features from the group of affinity matrix ${\textbf{\emph{H}}}_i$ and aggregated features ${\textbf{\emph{D}}}_i$. During the synthesis process, relevant spatial-temporal patterns should be enhanced while less relevant ones should be suppressed. To achieve that, a soft-attention map ${\textbf{\emph{G}}}$ is generated by taking the channel-wise maximum value on ${\textbf{\emph{H}}}$, which is obtained by concatenating a group of the affinity matrix ${\textbf{\emph{H}}}_i$ along the channel dimension.
In summary, the synthesized features $\widehat{\textbf{\emph{E}}}$ are calculated as follows:
\begin{equation}
\widehat{\textbf{\emph{E}}}=\textbf{\emph{E}}+Conv(\textbf{\emph{D}})\odot\textbf{\emph{G}},
\label{Inpainted}
\end{equation}
where $Conv$ and $\odot$ represent the convolutional and element-wise multiplication operation, respectively.
$\textbf{\emph{D}}$ is obtained by concatenating a group of the spatial-temporally aggregated features ${\textbf{\emph{D}}}_i$ along the channel dimension.

\subsubsection{Video Completion Network}
\label{Video Completion Network}
Video completion network aims to perceive valid context information from uncorrupted regions using predicted mask to generate corrupted contents, which is in line with the goal of video inpainting task in non-blind settings. 
Recently, benefiting from the advantages of long-range feature capture capacity, 
transformer-based video inpainting methods~\cite{cai2022devit,liu2021fuseformer,Ren_2022_CVPR,yan2020sttn,zhang2022flow}
have achieved unprecedented performance in non-blind settings.
These methods typically retrieve relevant contents to fill the corrupted regions by a self-attention mechanism.
However, they still suffer from two major limitations. 
Firstly, these methods always aggregate the features using all attention relations based on query-key pairs to generate the corrupted contents. 
Such a aggregation will leads to the redundant or irrelevant contents being filled into the corrupted regions, causing blurry or compromised results~\cite{chen2023learning,zhou2024adapt}. 
Secondly, these methods typically utilize the whole features to calculate self-attention.
Such a setting ignores the impact of noise in the features on attention calculation, significantly reducing the accuracy of attention retrieval~\cite{wu2024waveformer}.
Based on this, we design a wavelet sparse transformer as our video completion network to generate corrupted contents.

As shown in Fig.\ref{Fig_KT}(c), for the features $\textbf{\emph{F}}=\{\textbf{\emph{f}}_1, \textbf{\emph{f}}_2, \cdots,\textbf{\emph{f}}_T\}$ extracted by encoder, we first map each $\textbf{\emph{f}}_i\in\mathbb{R}^{{h}\times{{w}}\times{c}}$ into query ($\textbf{\emph{Q}}_i$), key ($\textbf{\emph{K}}_i$), and value ($\textbf{\emph{V}}_i$) to establish deep correspondences for each region in different semantic spaces. 
Having $\textbf{\emph{Q}}_i$, $\textbf{\emph{K}}_i$, and $\textbf{\emph{V}}_i$, we decompose them into different frequencies by Discrete Wavelet Transform (DWT)~\cite{1989A}.
Such a strategy can isolate the noise into the high-frequency components $\textbf{\emph{Q}}_i^H, \textbf{\emph{K}}_i^H, \textbf{\emph{V}}_i^H \in \mathbb{R}^{3\times{\frac{h}{2}}\times{{\frac{w}{2}}}\times{c}}$,  allowing the low-frequency components $\textbf{\emph{Q}}_i^L, \textbf{\emph{K}}_i^L, \textbf{\emph{V}}_i^L\in\mathbb{R}^{{\frac{h}{2}}\times{{\frac{w}{2}}}\times{c}}$ only contain relatively clean basic features.

After obtaining all low-frequency components, we use them to calculate standard dense self-attention (DSA):
\begin{equation}
DSA=Softmax\left(\frac{\textbf{\emph{Q}}^L\cdot{(\textbf{\emph{K}}^L)^T}}{\sqrt{d}}+\textbf{\emph{B}}\right),
\label{SSS}
\end{equation}
where $Softmax$ denotes the softmax layer. $\textbf{\emph{B}}$ refers to the learnable relative positional bias. Since not all query tokens are closely relevant to corresponding ones in keys, the utilization of all similarities is ineffective for generation corrupted contents. Instead, it leads to the redundant or irrelevant contents being filled into the corrupted regions, causing blurry or compromised results. 

Intuitively, developing a sparse self-attention (SSA) mechanism to aggregate the most relevant features could improve inpainting performance. 
Inspired by adaptive sparse self-attention~\cite{zhou2024adapt}, we utilize ReLu to remove negative similarities, preserving the most significant contents:
\begin{equation}
SSA=Softmax\left(ReLu\left(\frac{\textbf{\emph{Q}}^L\cdot{(\textbf{\emph{K}}^L)^T}}{\sqrt{d}}\right)+\textbf{\emph{B}}\right),
\label{SSS}
\end{equation}
where $ReLu$ denotes the ReLu layer. 
Note that simply using $SSA$ will impose over sparsity, making the learned feature representation insufficient to produce the complete corrupted contents. 
Conversely, using $DSA$ will inadvertently introduce irrelevant features into corrupted regions, leading to blurry results.
Therefore, we design a two-branch self-attention mechanism to calculate attention, maximizing the advantages of both two paradigms. 
The completed features of different frequencies can obtained by a shared attention manner:
\begin{equation}
\begin{split}
\widehat{\textbf{\emph{V}}}^L=&(\omega_1\odot DSA+\omega_2\odot SSA)\textbf{\emph{V}}^L,\\
\widehat{\textbf{\emph{V}}}^H=&(\omega_1\odot DSA+\omega_2\odot SSA)\textbf{\emph{V}}^H,
\label{SSS}
\end{split}
\end{equation}
where $\omega_1$ and $\omega_2$ are two normalized weights for adaptively modulating two-branch. 
$\widehat{\textbf{\emph{V}}}^L$ and $\widehat{\textbf{\emph{V}}}^H$ denote the completed low-frequency components and completed high-frequency components, respectively.  
In this way, the negative impact of noise on attention calculation can be significantly mitigated, thereby achieving the goal of improving the inpainting performance.
Finally, we utilize Inverse Discrete Wavelet Transform (IDWT) to obtain the final completed features $\widehat{\textbf{\emph{V}}}$.
Note that we only borrow information from valid regions to generate corrupted contents. 
Therefore, we set the $DSA$ and $SSA$ in corrupted regions to 0.

\subsection{Consistency Constraints}
\label{Consistency Constraints}
In our framework, mask prediction network and video completion network are closely correlated and mutually constrained.
The former helps the latter to locate the corrupted regions, while the latter regularizes the the former by the reconstruction loss, enforcing it to focus on corrupted regions. 

Ideally, if mask prediction network and video completion network both can capture accurate correspondence, the difference between the corrupted video frame $\textbf{\emph{x}}_i$ and the completed result $\widehat{\textbf{\emph{y}}}_i$ should exist only in the corrupted regions. 
This means that the following relationship holds:
\begin{equation}
\textbf{\emph{m}}_i=\mathcal{B}(\widehat{\textbf{\emph{y}}}_i-\textbf{\emph{x}}_i),
\label{STPM}
\end{equation}
\begin{equation}
\textbf{\emph{m}}_i^l=\mathcal{B}(\widehat{\textbf{\emph{y}}}_i-\textbf{\emph{x}}_i),
\label{STPM1}
\end{equation}
where $\mathcal{B}$ denotes binarization operation. $\textbf{\emph{m}}_i^{l}$ and $\textbf{\emph{m}}_i$ are the mask obtained by mask prediction network and ground truth mask, respectively.

Eq.(\ref{STPM}) and Eq.(\ref{STPM1}) give us the solution to further constraint mask prediction network and video completion network by a consistency loss. The consistency loss $\mathcal{{L}}_{c}$ is formulated as follows:
\begin{equation}
\mathcal{{L}}_{c}={\parallel{\textbf{\emph{m}}_i^l-\mathcal{B}(\widehat{\textbf{\emph{y}}}_i-\textbf{\emph{x}}_i)}\parallel_1+\parallel{\textbf{\emph{m}}_i-\mathcal{B}(\widehat{\textbf{\emph{y}}}_i-\textbf{\emph{x}}_i)}\parallel_1}.
\label{L-cycle_Y}
\end{equation}

\subsection{Loss Functions}
\label{Loss Functions}
We train our network by minimizing the following loss:
\begin{equation}
\begin{aligned}
\mathcal{{L}}=\lambda_{m}\mathcal{{L}}_{m}+\lambda_{v}\mathcal{{L}}_{v}+\lambda_{c}\mathcal{{L}}_{c},
\label{L-total}
\end{aligned}
\end{equation}
where $\mathcal{{L}}_{m}$, $\mathcal{{L}}_{v}$, and $\mathcal{{L}}_{c}$ denote
mask prediction loss, video completion loss and consistency loss, respectively. 
$\lambda_{m}$, $\lambda_{v}$ and $\lambda_{c}$ are non-negative trade-off parameters. 
In our experiments, 
$\lambda_{m}$, $\lambda_{v}$ and $\lambda_{c}$ are set to $3$, $5$ and $0.02$, determined by grid searching. More details of our framework and loss function can be found in the \textcolor{magenta}{supplementary materials}.

\begin{table*}[!t]
%\resize box
\caption{Quantitative results of video inpainting on YouTube-VOS and DAVIS datasets. The term $\emph{Blind}$ denotes `Blind setting' for short.}
  \centering
\small
    \begin{tabular}{c||c||c|c|c|c||c|c|c|c}
    \hline
    \hline
     ~   &   & \multicolumn{4}{c||}{\textbf{YouTube-VOS}}      & \multicolumn{4}{c}{\textbf{DAVIS}}  \\
     \cline{3-10}  \multirow{-2}{*}{ \textbf{Methods}} & \multirow{-2}{*}{ \textbf{Blind}}
       &  PSNR$\uparrow$ &  SSIM$\uparrow$ &   $E_{warp}\downarrow$  &  LPIPS$\downarrow$  &   PSNR$\uparrow$ &  SSIM$\uparrow$ &   $E_{warp}\downarrow$  &  LPIPS$\downarrow$  \\
    \hline
    \hline
    TCCDS~\cite{huang2016temporally} &\usym{1F5F4} &23.418&0.8119 &0.3388&1.9372 &28.146&0.8826   &0.2409&1.0079  \\
    \hline
    VINet~\cite{Kim_2019_CVPR,8931251} &\usym{1F5F4} &26.174 &0.8502 &0.1694 &1.0706 &29.149 &0.8965 &0.1846 &0.7262 \\
    \hline
    DFVI~\cite{xu2019deep}  &\usym{1F5F4} &28.672 &0.8706 &0.1479 &0.6285 &30.448&0.8961    &0.1640&0.6857 \\
    \hline
    FGVC~\cite{Gao-ECCV-FGVC} &\usym{1F5F4} &24.244&0.8114 &0.2484&1.5884&28.936&0.8852   &0.2122&0.9598 \\
    \hline
    CPVINet~\cite{lee2019cpnet} &\usym{1F5F4} &28.534    &0.8798&0.1613&0.8126&30.234&0.8997 &0.1892&0.6560  \\
    \hline
    STTN~\cite{yan2020sttn} &\usym{1F5F4} &28.993&0.8761 &0.1523&0.6965&28.891&0.8719   &0.1844&0.8683 \\
    \hline
    FuseFormer~\cite{liu2021fuseformer} &\usym{1F5F4} &29.765&0.8876 &0.1463&0.5481&29.627&0.8852&0.1767&0.6706 \\
    \hline 
    E2FGVI~\cite{li2022towards} &\usym{1F5F4} &30.064&0.9004&0.1490&0.5321  &31.941&0.9188&0.1579&{0.6344} \\
    \hline
    FGT~\cite{zhang2022flow} &\usym{1F5F4} &30.811  &0.9258&0.1308&0.4565  &32.742&0.9272 &0.1669&0.4240 \\
    \hline
     ProPainter~\cite{zhou2023propainter}   &\usym{1F5F4} &29.906    & 0.9050   &0.1458    & 0.4962   & 31.967   &0.9250    &0.1655    &0.4370  \\
     \hline
     WaveFormer~\cite{wu2024waveformer} &\usym{1F5F4} &{33.264}&{0.9435}&{0.1184}&{0.2933}&{34.169}&{0.9475}&{0.1504}&{0.3137} \\
     \hline
    \textbf{VCNet~(Ours)} &\usym{1F5F4} &\textbf{34.107}&\textbf{0.9521}&\textbf{0.1102}&\textbf{0.2145}&\textbf{34.936}&\textbf{0.9561}&\textbf{0.1362}&\textbf{0.2706} \\
    \hline
    \hline
    MPNet+CPVINet &\usym{2714}  &25.206   &0.8115   &0.2045   &1.2269   &26.024   &0.8296   &0.2178  &1.2543\\
    \hline
    MPNet+STTN &\usym{2714}  &25.049  &0.8295   &0.2032   &1.6490  &25.009   &0.8505   &0.2882  &1.4868\\
    \hline
    MPNet+FuseFormer &\usym{2714}  &25.269  &0.8302  &0.2062   &1.6772   &24.752  &0.8260  &0.2887  &1.4351\\
    \hline
    MPNet+E2FGVI &\usym{2714}  &26.422   &0.8368  &0.1832   &1.1145 & 27.318  &0.8536   &0.1871  &0.9523\\
    \hline
    MPNet+FGT &\usym{2714}  &27.032   &0.8755  &0.1609   &0.8667 & 28.387  &0.8963   &0.1759  &0.7392\\
    \hline
    MPNet+WaveFormer &\usym{2714}  &29.185   &0.8902  &0.1508   &0.7153 & 29.794  &0.9016   &0.1607  &0.7669\\
    \hline
    \textbf{Ours} &\usym{2714} &\textbf{30.528}&\textbf{0.9088}&\textbf{0.1362}&\textbf{0.6556}&\textbf{30.961}&\textbf{0.9107}&\textbf{0.1565}&\textbf{0.7338} \\
    \hline
    \hline
    \end{tabular}
  \label{tab_Q}
  \vspace{-0.2cm}
\end{table*}

\section{Experiments}
\label{Experiments}

\subsection{Dataset Customization}
\noindent\textbf{Training Dataset.}
Existing datasets~\cite{lee2019cpnet,chang2019free,yan2020sttn,Kang2022ErrorCF} usually generate corrupted video by ${\textbf{\emph{x}}_i}=(1-{\textbf{\emph{m}}_i})\odot{\textbf{\emph{y}}_i}+{\textbf{\emph{m}}_i}\odot{\textbf{\emph{u}}_i}$, where ${\textbf{\emph{u}}_i}$ is a pre-defined noise,  $\odot$ is the element-wise multiplication operation.
The non-blind setting always assume that the noise ${\textbf{\emph{u}}_i}$ is the “0" pixel value.
In this scenario, the corrupted regions $\textbf{\emph{m}}_i$ in corrupted video frame ${\textbf{\emph{x}}_i}$ can be easily identified and predicted by the prior knowledge introduced in the dataset, such as corrupted contents, clear border, and fixed shape.
Therefore, these datasets fail to realistically simulate the complex real-world scenarios in blind video inpainting.

In this paper, we customize a dataset suitable for blind video inpainting task. 
Specifically, we first utilize free-form strokes~\cite{chang2019free} as our corrupted regions ${\textbf{\emph{m}}_i}$, and fill ${\textbf{\emph{m}}_i}$ with real-world image patches as corrupted contents ${\textbf{\emph{u}}_i}$.
In this way, the generated corrupted contents have variable shapes, complex motions and faithful texture, effectively avoiding the introduction of shape and contents prior in the dataset.

After obtaining the corrupted regions ${\textbf{\emph{m}}_i}$ and corrupted contents ${\textbf{\emph{u}}_i}$, 
the corrupted video frame ${\textbf{\emph{x}}_i}$ can generated by ${\textbf{\emph{x}}_i}=(1-{\textbf{\emph{m}}_i})\odot{\textbf{\emph{y}}_i}+{\textbf{\emph{m}}_i}\odot{\textbf{\emph{u}}_i}$. 
However, the corrupted frame ${\textbf{\emph{x}}_i}$ obtained in this way often results in noticeable edges, which can serve as a strong indicator for identifying corrupted regions.
It may inadvertently encourage the model to learn and rely on this prior knowledge, ultimately compromising its ability to understand the semantic content of the video.
To avoiding the introduction of obvious edge priors, an iterative Gaussian smoothing~\cite{wang2018image} is introduce to extend the contact regions between ${\textbf{\emph{u}}_i}$ and ${\textbf{\emph{y}}_i}$. 
Such strategy can enforce the network to infer the corrupted regions by the semantic context of the video frame, rather than merely fitting to the training data.

Finally, to enhance the generalization ability of the model in practical applications, we also collect $1,250$ bullet removal video clips. 
We customized dataset consists of $2,400$ synthesized video clips and $1,250$ real-world video clips for blind video inpainting taks. 
This dataset will be published to facilitate subsequent research.

\noindent\textbf{Testing Dataset.}
To verify the effectiveness of the proposed blind video inpainting method, 
we employ two widely used datasets, namely Youtube-vos~\cite{Xu2018YouTube} and DAVIS~\cite{Perazzi2016A}, for evaluation. Following the previous works~\cite{Li2020ShortTermAL,xu2019deep}, $508$ and $60$ video clips of the Youtube-vos and the DAVIS are used as test sets to calculate metrics, respectively. Meanwhile, the corrupted contents of these test sets are synthesized using a similar manner to the training set. 

\subsection{Baselines and Metrics}
To the best of our knowledge, there is no work focusing on deep blind video inpainting task. Therefore, we use sixteen methods as our baselines to evaluate the blind video inpainting ability of our model, including eleven non-blind video inpainting methods, \emph{i.e.}, TCCDS~\cite{huang2016temporally}, VINet~\cite{Kim_2019_CVPR,8931251}, DFVI~\cite{xu2019deep}, CPVINet~\cite{lee2019cpnet}, FGVC~\cite{Gao-ECCV-FGVC}, STTN~\cite{yan2020sttn}, FuseFormer~\cite{liu2021fuseformer}, E2FGVI~\cite{li2022towards}, FGT~\cite{zhang2022flow}, ProPainter~\cite{zhou2023propainter}, and WaveFormer~\cite{wu2024waveformer}, and six blind video inpainting methods, \emph{i.e.}, MPNet+CPVINet~\cite{lee2019cpnet}, MPNet+STTN~\cite{yan2020sttn}, MPNet+FuseFormer~\cite{liu2021fuseformer}, MPNet+E2FGVI~\cite{li2022towards}, MPNet+FGT~\cite{zhang2022flow}, and MPNet+WaveFormer~\cite{wu2024waveformer}.
To ensure the fairness of the experimental results, non-blind video inpainting baselines follow their original non-blind settings (including model input and test datasets), while blind video inpainting baselines are fine-tuned on our synthesized dataset using their released models and codes.
Similar to the previous works~\cite{9967838,9446636}, we employ four metrics to report quantitative results, \emph{i.e.}, PSNR~\cite{9008384}, SSIM~\cite{9010390}, LPIPS~\cite{zhang2018unreasonable} and flow warping error $E_{warp}$~\cite{lai2018learning}.

\begin{figure}[tb]
\centering%height=3.0cm,width=15.5cm
\includegraphics[scale=0.67]{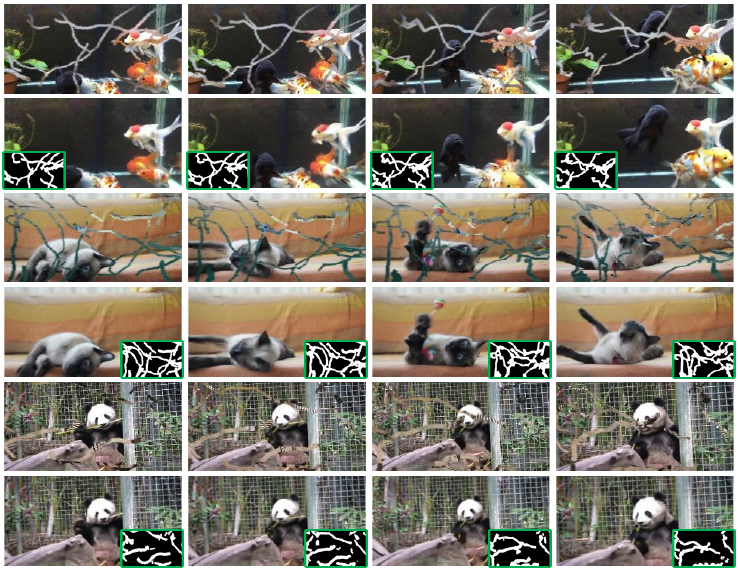}
\vspace{-0.5cm}
\caption{Three example of inpainting results with our method. The top row shows corrupted video frame. The completed results are shown in the bottom row, where green box denotes the mask generated by the model.}
\label{Fig6}
\end{figure}

\begin{figure}[tb]
\centering%height=3.8cm,width=17.5cmscale=0.75
\includegraphics[scale=0.67]{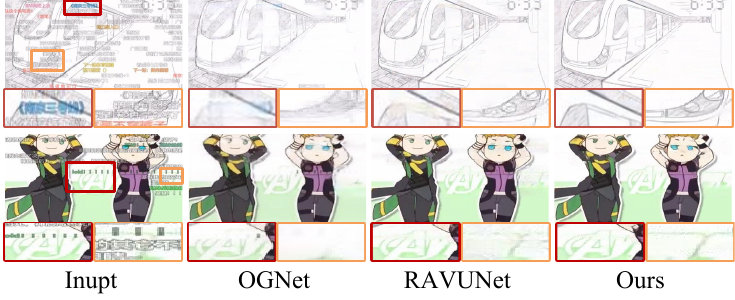}
\vspace{-0.6cm}
\caption{Qualitative results compared with OGNet~\cite{phutke2023blind} and RAVUNet~\cite{agnolucci2022restoration} on bullet removal.}
\label{Fig_BIRC}
\vspace{-0.2cm}
\end{figure}

\subsection{Experimental Results on Synthesized Dataset}
We report the quantitative evaluation results of our method and other baselines in Tab.~\ref{tab_Q}. As shown in this table, the PSNR, SSIM, $E_{warp}$ and LPIPS of our method achieve state-of-the-art results under blind inpainting settings on two datasets, and have obtained comparable performance with non-blind video inpainting methods. In particular, for blind inpainting settings, our method outperforms the best baseline (MPNet+FGT~\cite{zhang2022flow}) with a large margin in all evaluation metrics. 
Compared with video inpainting methods under non-blind settings, our blind inpainting model achieves a comparable performance to the sub-optimal baseline E2FGVI~\cite{li2022towards}. 
This verifies the superiority and effectiveness of the proposed blind video inpainting approach.
To further compare the visual qualities of inpainted results,
we show the inpainted results for our model on three examples in Fig.~\ref{Fig6}. 
As can be observed, our method can obtain spatial-temporally consistent inpainted results without any mask annotations. 
This demonstrates the effectiveness of the proposed method. 
More blind video inpainting results can be found in the \textcolor{magenta}{supplementary materials}.

\subsection{Experimental Results on Real Cases}
The proposed blind video inpainting method is beneficial for many practical applications, such as bullet removal. Fig·~\ref{Fig_BIRC} compares the results of bullet removal between OGNet~\cite{phutke2023blind}, RAVUNet~\cite{agnolucci2022restoration}, and our method. As shown in Fig.~\ref{Fig_BIRC}, our method effectively eliminates bullets in videos without the need for any mask annotations, and generates better details than the baselines. This results further demonstrate the effectiveness and superiority of our method in practical applications. More real cases can be found in the \textcolor{magenta}{supplementary materials}.

\subsection{Ablation Studies}
\label{Ablation Studies}
\begin{table}[tb]
\footnotesize
\renewcommand\tabcolsep{3.5pt}
  \centering
  % \vspace{-0.2cm}
  \caption{Ablation study of MPNet.}
  \vspace{-0.2cm}
    \begin{tabular}{c|cc||c|cc}
    \hline
    \hline
     \rowcolor{mygray} \textbf{Methods} &BCE$\downarrow$ & IOU$\uparrow$   & \textbf{Methods} &BCE$\downarrow$ & IOU$\uparrow$  \\
    \hline
    \hline
    STP &1.1251       &0.8437      & DWT\_STP &1.0785  &0.8682  \\
    DWT\_STP+LTR &0.9176    &0.8829       & Full MPNet &\textbf{0.8052}       &\textbf{0.9017}  \\
    \hline
    \hline
    \end{tabular}%
  \label{tab_M}%
  \vspace{-0.3cm}
\end{table}%
\begin{figure}[tb]
\centering%height=3.0cm,width=15.5cm
\includegraphics[scale=0.67]{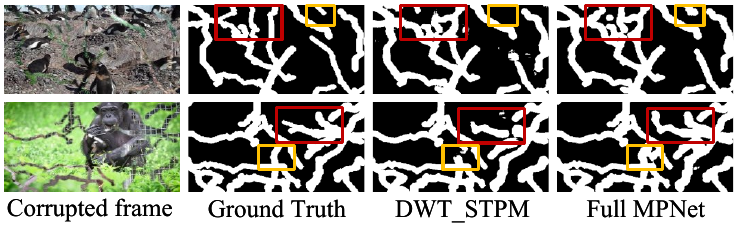}
 \vspace{-0.6cm}
\caption{Example of corrupted regions segmentation.}
\label{Fig_MV}
\vspace{-0.1cm}
\end{figure}
\begin{figure}[!t]
\centering%height=3.8cm,width=17.5cmscale=0.75
\includegraphics[scale=0.67]{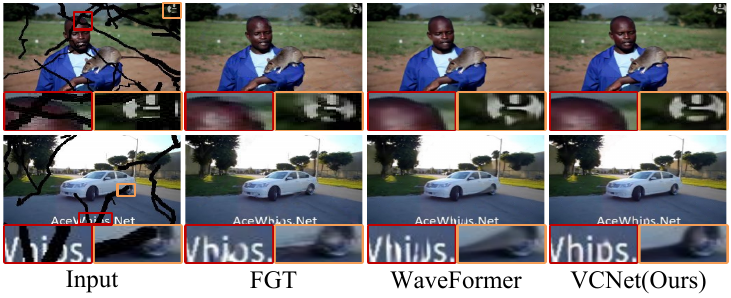}
\vspace{-0.7cm}
\caption{Comparison of qualitative results under non-blind settings. Better viewed at zoom level 400\%.}
\vspace{-0.2cm}
\label{Fig_VQ}
\end{figure}

\begin{figure*}[!t]
\centering%height=3.8cm,width=17.5cmscale=0.75
\includegraphics[scale=0.708]{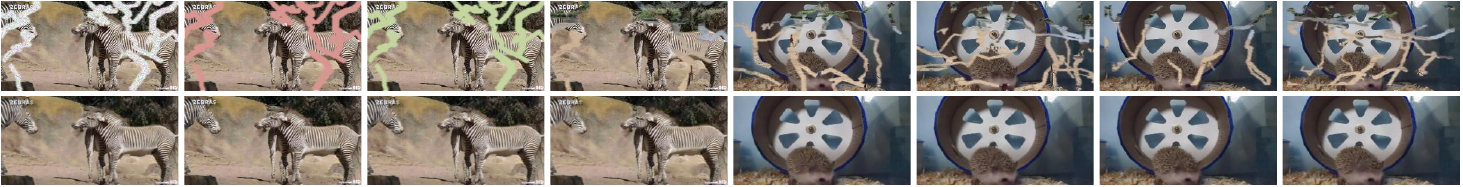}
\vspace{-0.5cm}
\caption{(a) Visual evaluations with random masks filled with different content (first four examples). (b) Visual comparison of different masks filled with the same content (last four examples). Better viewed at zoom level 400\%.}
\label{Fig_RCM}
\vspace{-0.2cm}
\end{figure*}

\noindent\textbf{Effectiveness of MPNet.}
We conduct an ablation study on the proposed MPNet. 
As shown in Tab.~\ref{tab_M}, we compare the full MPNet model with its three variants,
\emph{i.e.}, “STPM”, “DWT\_STPM”, and “DWT\_STPM+LTRM”.
First two variants refer to down-sampling in the short-term prediction module using the strided-convolution and the DWT, separately.
The third variant refines the prediction result of the second variant using strided-convolution for down-sampling in the long-term refinement module.
The results indicate that the full MPNet model achieves the best prediction performance of corrupted regions, 
with the values of 0.8052 and 0.9017 for binary cross entropy (BCE) and intersection over union (IOU), respectively. 
Besides, we also show the predicted results of the full MPNet model and the second variant. 
As revealed in Fig.~\ref{Fig_MV}, the corrupted regions predicted by the full MPNet model are closer to ground-truth. This demonstrates that the effectiveness of MPNet.

\noindent\textbf{Effectiveness of VCNet.}
To demonstrate the effectiveness of VCNet, we compare it with ten non-blind video inpainting methods.
As shown in Tab.~\ref{tab_Q}, our VCNet achieves substantial improvements on all four quantitative metrics.
Furthermore, we also display some qualitative comparisons with two representative methods, including FGT~\cite{zhang2022flow} and WaveFormer~\cite{wu2024waveformer}. As shown in Fig.~\ref{Fig_VQ}, our VCNet can generate faithful texture and structure information, which verifies the effectiveness and superiority of VCNet.

\noindent\textbf{Effectiveness of Sparse Self-Attention.}
In Tab.~\ref{tab_msfn}, we perform an ablation study for sparse self-attention. From the table, we can observe that: 1) The model using only the dense self-attention (DSA) branch achieves the worst inpainting performance due to the introduction of irrelevant features. 2) Using sparse self-attention (SSA) branches improves the inpainted performance of the model. 
These results further demonstrates the necessity of an adaptive selection strategy for video inpainting task.

\noindent\textbf{Effectiveness of Consistency Loss $\mathcal{{L}}_{c}$.}
$\mathcal{{L}}_{c}$ is introduced to regularize the training parameters, enforcing mask prediction network and video completion network to capture accurate correspondence. Tab.~\ref{tab_msfn} presents an ablation study on $\mathcal{{L}}_{c}$. As shown in Tab.~\ref{tab_msfn}, the models with $\mathcal{{L}}_{c}$ participating in training can obtain better inpainted results.
\begin{table}[!t]
\vspace{-0.2cm}
\footnotesize
\centering
\caption{Ablation study of sparse self-attention and $\mathcal{{L}}_{c}$.}
\vspace{-0.2cm}
\begin{tabular}{ccc||cccc}
\hline
\hline
 \rowcolor{mygray} \multirow{1}{*}{DSA}                            & 
\multirow{1}{*}{SSA}                            & 
\multirow{1}{*}{$\mathcal{{L}}_{c}$}                            & 
\multirow{1}{*}{PSNR$\uparrow$} & 
\multirow{1}{*}{SSIM$\uparrow$} & 
\multirow{1}{*}{$E_{warp}\downarrow$} & 
\multirow{1}{*}{LPIPS$\downarrow$} \\ 
\hline
\hline
\usym{2714} & 
\multicolumn{1}{c}{} &  
&29.172       &0.8897      &0.1529       & 0.7264 \\ 
\usym{2714}  & 
\usym{2714} &  
&29.885      &0.8962      & 0.1454      &0.6891  \\ %\hline
\usym{2714}  & 
\usym{2714}  & \usym{2714} 
 &\textbf{30.528}&\textbf{0.9088}&\textbf{0.1362}&\textbf{0.6556}\\ 
\hline
\hline
\end{tabular}
\vspace{-0.3cm}
\label{tab_msfn}
\end{table}

\noindent\textbf{Robustness against Various Degradation Patterns.}
In Fig.~\ref{Fig_RCM}(a), we filled the corrupted regions of the video with different contents to perform the inpainting experiment. As shown in figure, our model can handle Gaussian noise or constant color fills, even though these patterns are not included in the training dataset. 
These results suggest that our model can learn to identify and inpaint the visually inconsistent regions in video frames, rather than simply memorizing the data distribution of the synthetic dataset.

\noindent\textbf{Robustness against Various Corrupted Patterns.}
We conduct an ablation study for corrupted patterns. As illustrated in Fig.~\ref{Fig_RCM}(b), our blind video inpainting model can effectively handle corrupted regions of various patterns, and generates visually pleasing results. This indicates that our method is highly robust to the patterns of corrupted regions.

\begin{table}[!t]
\footnotesize
\renewcommand\tabcolsep{0.5pt}
  \centering
  \caption{Efficiency analysis.}
  \vspace{-0.2cm}
    \begin{tabular}{c||cccccc}
    \hline
    \hline 
    \rowcolor{mygray} {Metrics}  & STTN~\cite{yan2020sttn}  &FuseFormer~\cite{liu2021fuseformer}   & E2FGVI~\cite{li2022towards}  &FGT~\cite{zhang2022flow} & VCNet(Ours)\\
    \hline
    \hline
    FLOPs &477.91G &579.82G   &442.18G   & 455.91G  &\textbf{396.35G} \\
    Time  &0.22s   &0.30s     &0.26s &0.39s   &\textbf{0.21s}\\
    \hline
    \hline
    \end{tabular}%
    \vspace{-0.3cm}
  \label{AS_ETC}%
\end{table}%

\noindent\textbf{Efficiency Analysis.}
We compare the efficiency of our VCNet with STTN~\cite{yan2020sttn}, FuseFormer~\cite{liu2021fuseformer}, E2FGVI~\cite{li2022towards} and FGT~\cite{zhang2022flow} on two metrics: FLOPs and inference time. 
Following previous works \cite{liu2021fuseformer,yan2020sttn,zhang2022flow}, we measure FLOPs with a temporal size (number of frames) of 20.
The inference time is measured on a single Titan RTX. 
As shown in Tab.~\ref{AS_ETC}, our VCNet achieves the fastest inference speed and lowest FLOPs compared to baselines. 
This also further indicates that our full model (MPNet+VCNet) is more efficient than MPNet+STTN~\cite{yan2020sttn}, MPNet+FuseFormer~\cite{liu2021fuseformer}, MPNet+E2FGVI~\cite{li2022towards}, and MPNet+FGT~\cite{zhang2022flow}. 

%More ablation experiments can be found in the \textcolor{magenta}{supplementary materials}.

\section{Conclusion}
\label{Conclusion}
In this paper, we formulate a novel blind video inpainting task,
which aims to achieve video inpainting without any corrupted regions annotation.
For this purpose, we design an end-to-end blind video inpainting framework consisting of a mask prediction network and a video completion network, and a consistency loss is introduced to regularize the training parameters.
Furthermore, we customize a dataset suitable for blind video inpainting, which can effectively facilitate subsequent researches.
Experimental results demonstrate the effectiveness of the our method.

{   
\small
    \bibliographystyle{ieeenat_fullname}
    \bibliography{main}
}


\end{document}